\documentclass[letterpaper, 10 pt, journal, twoside]{ieeetran}

\IEEEoverridecommandlockouts                             
\usepackage{graphicx}
\usepackage{marvosym}
\usepackage{xcolor} 
\usepackage[svgnames]{xcolor}
\usepackage[colorlinks=true, linkcolor=LightSlateGrey, citecolor=LightSlateGrey, urlcolor=LightSlateGrey]{hyperref}
\usepackage{booktabs}
\usepackage{multirow}
\usepackage{amsmath}
\usepackage{amssymb}
\usepackage[style=ieee]{biblatex}
\usepackage{subfiles} 

\newif\ifhighlight
\highlightfalse

\newcommand{\hltext}[1]{%
  \ifhighlight
    \textcolor{blue}{#1}
  \else
    #1
  \fi
}

\begin{document}

\title{
VLM-SFD: VLM-Assisted Siamese Flow Diffusion Framework for Dual-Arm Cooperative Manipulation
}

\author{Jiaming Chen$^{*}$$^{1}$, Yiyu Jiang$^{*}$$^{1}$, Aoshen Huang$^{2}$, Yang Li$^{\text{\Letter}1}$, and Wei Pan$^{1}$%
\thanks{Received 2 June 2025; accepted 13 October 2025. This article was recommended for publication by Associate Editor C. Jara and Editor J. Borràs Sol upon evaluation of the reviewers' comments.
The work of Jiaming Chen is supported by The University of Manchester-China Scholarship Council Joint Scholarship. (* Jiaming Chen and Yiyu Jiang contributed equally to this work.) ($^{\text{\Letter}}$ Corresponding author: Yang Li.)}  
\thanks{$^{1}$Jiaming Chen, Yiyu Jiang, Yang Li, and Wei Pan are with the Department of Computer Science, The University of Manchester, Manchester M13
9PL, United Kingdom.
        {(email: \tt\footnotesize ppjmchen@gmail.com; yiyu.jiang@postgrad.manchester.ac.uk; yang.li-4@manchester.ac.uk; wei.pan@manchester.ac.uk})}%
\thanks{$^{2} $Aoshen Huang is with the School of Control Science and Engineering,
Shandong University, Jinan 250061, China (e-mail: \tt\footnotesize aoshenhuang @mail.sdu.edu.cn).}%
\thanks{This article has supplementary downloadable material available at ~\url{https://doi.org/10.1109/LRA.2025.3627381}, provided by the authors.}
\thanks{Digital Object Identifier (DOI): 10.1109/LRA.2025.3627381}
}

\maketitle

\markboth{IEEE Robotics and Automation Letters. Preprint Version. Accepted October, 2025}
{Chen \MakeLowercase{\textit{et al.}}: VLM-SFD: VLM-Assisted Siamese Flow Diffusion Framework for Dual-Arm Cooperative Manipulation} 

\begin{abstract}
Dual-arm cooperative manipulation holds great promise for tackling complex real-world tasks that demand seamless coordination and adaptive dynamics.
Despite substantial progress in learning-based motion planning, most approaches struggle to generalize across diverse manipulation tasks and adapt to dynamic, unstructured environments, particularly in scenarios involving interactions between two objects such as assembly, tool use, and bimanual grasping.
To address these challenges, we introduce a novel VLM-Assisted Siamese Flow Diffusion (VLM-SFD) framework for efficient imitation learning in dual-arm cooperative manipulation. 
The proposed VLM-SFD framework exhibits outstanding adaptability, significantly enhancing the ability to rapidly adapt and generalize to diverse real-world tasks from only a minimal number of human demonstrations.
Specifically, we propose a Siamese Flow Diffusion Network (SFDNet) employs a dual-encoder-decoder Siamese architecture to embed two target objects into a shared latent space, while a diffusion-based conditioning process—conditioned by task instructions—generates two-stream object-centric motion flows that guide dual-arm coordination. 
We further design a dynamic task assignment strategy that seamlessly maps the predicted 2D motion flows into 3D space and incorporates a pre-trained vision-language model (VLM) to adaptively assign the optimal motion to each robotic arm over time.
Experiments validate the effectiveness of the proposed method, demonstrating its ability to generalize to diverse manipulation tasks while maintaining high efficiency and adaptability. 
The code and demo videos are publicly available on our project website~\url{https://sites.google.com/view/vlm-sfd/}.


\end{abstract}

\section{Introduction}

\IEEEPARstart{D}{ual-arm} cooperative manipulation involves the synchronized control of two robotic arms to execute complex tasks with high precision and adaptability~\cite{he2022double,kruger2011dual,bai2022anthropomorphic}. Compared to the extensively-studied field of single-arm manipulation~\cite{im2flow2act,chi2024universal}, dual-arm manipulation poses significant challenges in coordination and control, requiring precise synchronization, collision avoidance, and adaptive planning~\cite{dastider2024apex}.


Previous research has explored various directions, including control optimization~\cite{ibvs,tarbouriech2018dual}, vision-based learning~\cite{zhao2023learning, liu2024rdt}, and advanced motion planning frameworks~\cite{sintov2020motion, yu2023coarse}. While these methods enhance adaptability and synchronization, they often rely on extensive dual-arm demonstrations and require meticulous tuning, limiting their generalization capabilities. More recently, flow-based policies~\cite{orion,im2flow2act} and diffusion models~\cite{im2flow2act} have shown promise in robotic manipulation, demonstrating improved motion prediction while requiring fewer human demonstrations. 
Despite these advancements, extending flow-based methods to dual-arm cooperative manipulation presents significant yet largely unaddressed research challenges including modeling the interactions between two manipulators, ensuring synchronized coordination while avoiding collisions, task allocations between manipulators based on spatial and temporal constraints, and lacking of sufficient real-world data of dual-arm demonstrations.

To overcome these challenges, we introduce the \textit{VLM-Assisted Siamese Flow Diffusion} (VLM-SFD) framework. This framework enables the learning of coupled dual-arm behaviors from a small set of human demonstrations by leveraging the proposed \textit{Siamese Flow Diffusion Network} (SFDNet) and a novel VLM-assisted spatial-temporal task allocation strategy.
The SFDNet employs a dual-encoder-decoder architecture to generate object-centric motion flows through a conditional diffusion process. 
Initially, target objects are localized and extracted motion flows which are then encoded by a Siamese Variational Autoencoder (VAE) encoder, yielding latent features that capture object properties and interactions. 
These latent features are then fed into a Siamese UNet, conditioned on the task instruction, followed by a Siamese VAE decoder which iteratively refines the motion flows through diffusion, generating two synchronized and compatible motion streams that guides dual-arm coordination.

With these object-centric motion flows, it is crucial to assign the suitable motion trajectory to each arm while considering both spatial feasibility and temporal order for collision avoidance and efficient execution. 
To achieve this, we introduce a VLM-Assisted Spatial-Temporal Task Allocation Strategy, leveraging a pre-trained vision-language model (VLM) for knowledge-driven reasoning. Given a task instruction and visual clues, the VLM interprets task semantics, object relationships, and execution dependencies, dynamically assigning motion flows to each manipulator over time. This VLM-guided allocation prevents collisions and ensures sequential, efficient execution, allowing direct deployment on real-world robotic platforms without additional fine-tuning or real-world data.

We evaluated the VLM-SFD framework on four challenging real-world tasks, demonstrating its data efficiency and deployment advantages. Unlike prior imitation learning-based methods which often require extensive data and real-world fine-tuning, VLM-SFD achieves competitive performance with only ten human demonstrations for each task. Furthermore, our model can be directly deployed on a physical dual-arm robotic platform without further fine-tuning. Experimental results confirm that VLM-SFD maintains high success rates and efficient execution, despite using significantly fewer training examples and requiring no extra real-world adaptation, underscoring its advantage in practical deployment scenarios.



The contributions of this work are summarized as follows:
\begin{itemize}
\item We propose VLM-SFD, a novel VLM-Assisted Siamese Flow Diffusion framework that integrates two-stream motion flow generation and spatial-temporal task allocation for dual-arm cooperative manipulation. 
This framework enables rapid adaptation and deployment to new real-world tasks using a small set of demonstrations.
\item We introduce SFDNet, a dual-encoder-decoder Siamese Flow Diffusion Network that applies conditional diffusion for dual-arm action generation. To the best of our knowledge, this is the first framework to leverage diffusion-based motion synthesis in a dual-arm setting, \hltext{enabling flexible coordination and synchronized or sequential behaviors from limited demonstration data.}
\item We design a VLM-assisted spatial-temporal task allocation strategy, incorporating spatial-temporal knowledge reasoning to dynamically assign motion flows to each manipulator. By leveraging a pre-trained vision-language model, our method interprets task semantics and scene context to optimize execution order and prevent conflicts.
\item The integrated VLM-SFD framework is validated through real-world experiments on four challenging tasks, demonstrating superior success rates and execution efficiency compared to baseline methods, confirming its effectiveness in practical deployment scenarios.
\end{itemize}

The remainder of this paper is structured as follows: Section~\ref{sec:related_work} reviews related work on dual-arm cooperative manipulation and flow-based policy. The proposed VLM-SFD framework is then introduced in Section~\ref{sec:method}. Section~\ref{sec:exp} details the experimental setup and presents the corresponding results. Finally, the conclusion and limitations are discussed in Section~\ref{sec:conclusion}.



\section{Related Work}
\label{sec:related_work}
\subsection{Dual-Arm Cooperative Manipulation}

Dual-arm cooperative manipulation involves two robotic arms working together through synchronized planning and task-space coordination for effective object manipulation.
Previous work in this area has explored methods~\cite{jiao2022adaptive, zhao2023learning, yu2023coarse} from low-level control to high-level task execution.
One major research direction focuses on the optimization of control strategies, including dynamical system modeling~\cite{ren2024enabling}, hybrid control approaches~\cite{jiao2022adaptive, ren2017adaptive}, and adaptive force interaction~\cite{lu2018fuzzy}. These methods enable precise force modulation and trajectory execution, making them particularly relevant for high-precision tasks such as robot-assisted surgery~\cite{wu2019coordinated}, cloth manipulation~\cite{liu2024rdt}, and assembly~\cite{wang2024dexcap}. \textcolor{black}{While computer vision has advanced significantly in robotics, another emerging direction is vision-based learning, particularly through reinforcement learning and imitation learning, which making robots more intelligent and adaptable to complex tasks and environments~\cite{soori2023artificial}.}
Methods like DexCap~\cite{wang2024dexcap} and ALOHA~\cite{zhao2023learning} leverage teleoperation for efficient data collection of bimanual skills, while RDT-1B~\cite{liu2024rdt} and Bi-DexHands~\cite{chen2023bi} employ large-scale bimanual training datasets to improve dexterity across diverse scenarios. Although these methods offer improved scene adaptability and skill transfer, they often rely on extensive data collection to generalize effectively. 
\textcolor{black}{In addition to learning-based approaches, traditional control methods like sparse kinematic control~\cite{tarbouriech2018dual} and IBVS~\cite{ibvs} have been explored for dual-arm cooperation. Sparse kinematic control minimizes the number of actuated joints for efficient coordination, while IBVS integrates visual feedback to adjust manipulator poses, reducing synchronization errors. However, these methods are often task-specific and struggle to generalize across diverse scenarios.} Recently, various motion planning frameworks have been developed for high-level task execution in dual-arm cooperative manipulation. Yu et al.~\cite{yu2023coarse} and Sun et al.~\cite{sun2022motion} both explored hierarchical motion planning to handle complex tasks with dynamic obstacle avoidance. While these methods improve adaptability and smoothness, they still struggle to generalize to unstructured environments and diverse tasks.

\subsection{Flow-Based Policy}

Flow-based policies are a promising technique for robotic manipulation, emphasizing the integration of object flows to enhance performance in complex manipulation tasks. Previous research on object-centric manipulation has attempted to extract 3D point cloud flow representations for skill learning in domain transfer scenarios~\cite{orion}. However, these approaches have faced limitations in scalability and generalization. With advancements in computer vision, recent works such as Track2Act~\cite{bharadhwaj2024track2act} and ATM~\cite{wen2023any} have applied flow-based tracking prediction from video datasets to guide skill learning for general robotic manipulation tasks. \textcolor{black}{Other studies have further extended flow-based policies to cross-domain skill transfer~\cite{xu2024flow}, language-conditioned 3D flow prediction~\cite{yuan2024general}, and improved inference efficiency for flow-based policies~\cite{zhang2024flowpolicy}.} These approaches primarily focus on interaction between the manipulator and target objects, enhancing skill transferability, and improving generalization across objects and different scenerioes. However, the interaction between objects remain significant challenge, particularly in unstructured environments. \textcolor{black}{To address this, Weng et al.~\cite{weng2022fabricflownet} introduced FabricFlowNet (FFN) for bimanual cloth manipulation, leveraging optical flow to improve performance. Although FFN demonstrates generalization across different cloth shapes, it remains constrained in its ability to generalize beyond fabric manipulation and does not explicitly model object-object interactions.}

\section{Method}
\label{sec:method}
\begin{figure*}[ht]
    \centering
    \includegraphics[width=0.9\textwidth]{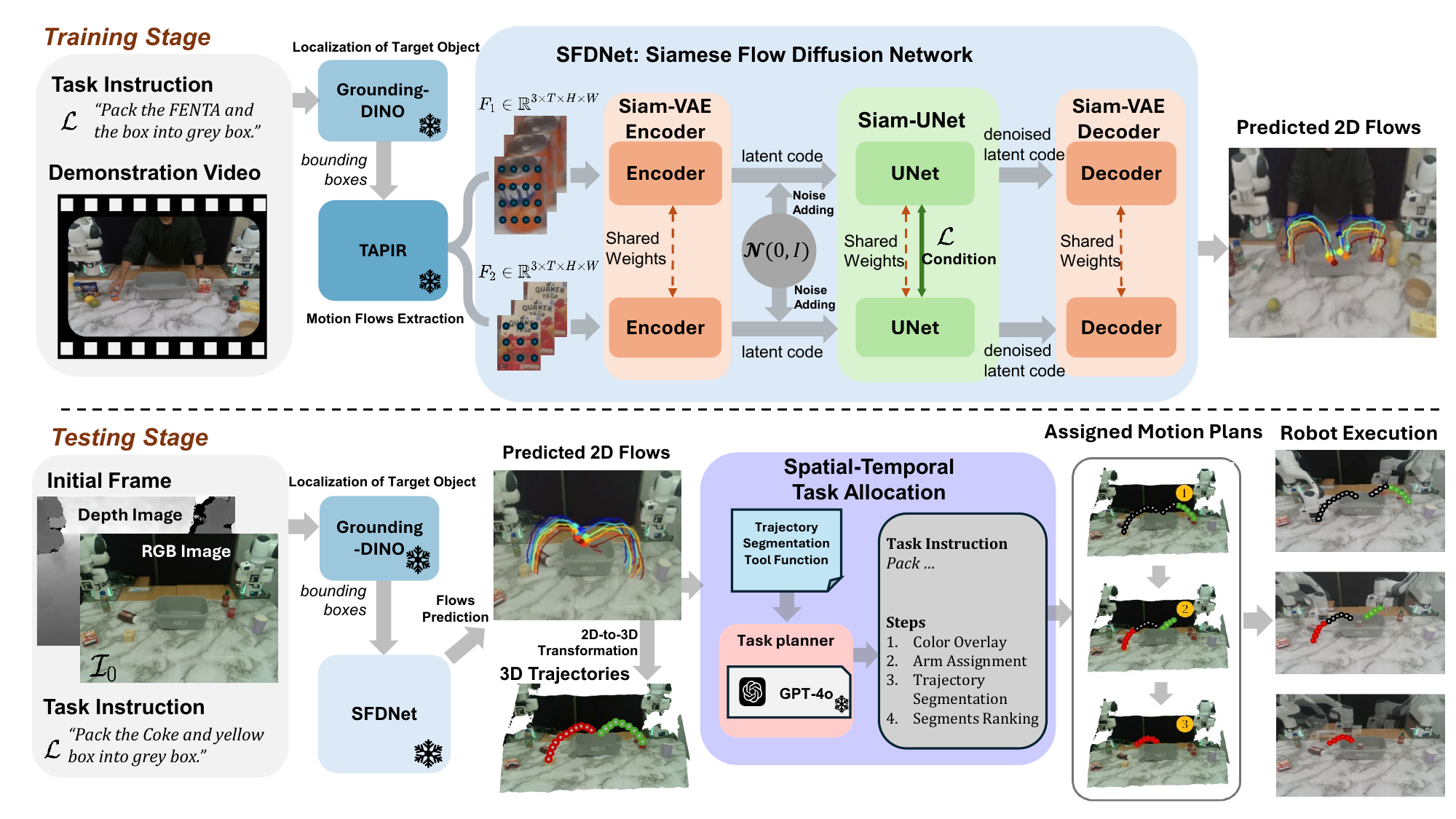}
    \caption{Overview of VLM-SFD Framework. In the training stage, demonstration video and task instruction are first fed into Grounding-DINO~\cite{dino} and TAPIR~\cite{tapir} for localizing the objects to be manipulated and extracting corresponding flow, which are then used for training the Siamese Flow Diffusion Network (SFDNet) to predict dense 2D motion flows. In the testing stage, given an initial frame and an instruction, the model predicts 2D flows, which are then post-processed into 3D trajectories, which are further fed into the Spatial-Temporal Task Allocation module to generate coordinated dual-arm motion plans for robot execution.
    }
    \label{fig_method}
\end{figure*}

\subsection{Overview: VLM-SFD Framework}

\textcolor{black}{Dual-arm cooperative manipulation requires two robotic arms to work synergistically on tasks involving multiple objects, enabling complex \textcolor{black}{cooperative} manipulation.}
To address this challenge, we propose a VLM-assisted Siamese flow diffusion framework, which formulates the problem as a two-stage process, as illustrated in Fig.~\ref{fig_method}.
In the training stage, the system learns object-centric motion flows for the objects involved in the interaction. In testing stage, the learned 2D motion flows are mapped into the 3D execution space, ensuring that both arms move in precise coordination.

In the training stage, we collect a small set of human demonstration videos as training data for generating motion flows in dual-arm cooperative manipulation.
Inspired by Im2Flow2Act~\cite{im2flow2act}, we take the initial frame $\mathcal{I}_0$ and a natural language task instruction $\mathcal{L}$ as inputs. 
To facilitate object identification and localization, particularly crucial for dual-arm cooperative tasks involving two target objects, we employ the pre-trained 
Grounding-DINO~\cite{dino}, represented by $\mathbf{G}$. 
This module effectively maps the language instruction to the visual scene, enabling the precise localization of relevant objects. 
Mathematically, this process can be formalized as:
\begin{equation}
    \{O_1, O_2\} = \mathbf{G}(\mathcal{I}_0, \mathcal{L}|\phi_\mathbf{G}), 
\end{equation}
where $\phi_\mathbf{G}$ denotes the pre-defined prompt for visual grounding, and $O_1\in \mathbb{R}^4,O_2\in \mathbb{R}^4$ are the bounding boxes of the target objects.

Subsequently, we utilize TAPIR~\cite{tapir} as a visual tracking module, denoted as $\mathcal{T}$. 
By providing the target object bounding boxes $O_1, O_2$ as queries, the tracker generates two-stream object-centric motion flows $F_1$ and $F_2$ from the demonstration video $\mathcal{V}$. The process could be formulated as: 
\begin{equation}
    \{F_1, F_2\}=\mathcal{T}(\mathcal{V}|O_1, O_2),
\end{equation}
where $F_1\in \mathbb{R}^{3\times T\times H\times W}, F_2\in \mathbb{R}^{3\times T\times H\times W}$ is the extracted motion flows, \textcolor{black}{where the first two dimensions are 2D coordinates of each point within flow and the last dimension notes the visibility of the point}, $T$ denotes the temporal length of the video, and $H, W$ are the spatial dimensions of the extracted motion flow. 
Each motion flow encodes the per-pixel displacement of the target objects over time, offering a structured representation of their movement within the scene and serving as supervision for training the SFDNet.


In the testing stage, we deploy the trained SFDNet directly onto a real-world dual-arm platform. The predicted 2D object-centric motion flows are first mapped into the 3D execution space. By leveraging VLM-assisted spatial-temporal knowledge reasoning, the framework dynamically interprets task semantics, object relationships, and execution dependencies to assign motion trajectories to each manipulator. This enables adaptive scheduling and collision-free coordination, optimizing the execution order based on spatial constraints and temporal dependencies within the task.

The details of the SFDNet and the VLM-assisted task assignment strategy are introduced in the following sections.



\textcolor{black}{
\subsection{SFDNet: Siamese Flow Diffusion Network} 
}


As illustrated in Fig.~\ref{fig_method}, SFDNet consists of three core modules, a Siam-VAE Encoder, a Siam-UNet, and a Siam-VAE Decoder that collectively capture high-level task context while generating fine-grained motion flows.

\paragraph{Siam-VAE Encoder}

\textcolor{black}{
We design a Siamese-structured Variational Autoencoder encoder (Siam-VAE Encoder) with two parallel branches that share weights. Each branch takes the object-centric motion flow data for one target as the input, thereby embedding both flows into a unified latent space. Formally, each branch first produces a latent code sequence:
\begin{equation} 
\mathbf{z}^{1:T}_i(0) =\mathcal{E}\bigl(F_i\bigr), \quad i \in \{1,2\}, 
\end{equation}
where $\mathbf{z}^{1:T}_i(0)$ denotes the initial latent code, which serves as the input for the first step of the diffusion process}, $\mathcal{E}(\cdot)$ is the pre-trained VAE encoder that encodes the flow frame-wisely. 
Since both branches share parameters, the model learns consistent spatiotemporal representations across objects while preserving object-specific characteristics. The two latent code sequences $\mathbf{z}^{1:T}_1(0)$ and $\mathbf{z}^{1:T}_2(0)$ then serve as inputs of the following diffusion process.



\paragraph{Siam-UNet} 
We then design a Siamese-structured UNet (Siam-UNet) to perform the diffusion process for learning two-stream object-centric motion flow generation.  Inspired by AnimateDiff~\cite{animatediff} and Im2Flow2Act~\cite{im2flow2act}, the latent codes $\mathbf{z}^{1:T}_i(0)$ are perturbed by the forward diffusion:
\begin{equation}
\mathbf{z}^{1:T}_i(t)=\sqrt{\bar{\alpha}(t)}\mathbf{z}^{1:T}_i(0) + \sqrt{1-\bar{\alpha}(t)}\epsilon^{1:T}_i, 
\label{eq_forward}
\end{equation}
where $t$ denote the diffusion forward step, $\bar{\alpha}(t)$ denote the pre-defined noise strength at step $t$, and $\epsilon^{1:T}_i$ is an auxiliary variable sampled from Gaussian noise $\mathcal{N}(0,I)$ in each branch. 


To encode the task instruction as a condition for the following diffusion process, the multi-head cross-attention (MHCA) mechanism~\cite{attention} is employed to perform information exchange between latent codes and instruction. For conciseness, the MHCA are formally defined as follows:
\begin{equation} 
\label{MHCA} 
\text{MHCA}(X, Y) = \sigma(f_{Q}(X)f_{K}(Y)^\mathsf{T})f_{V}(Y), 
\end{equation}
where $\sigma$ is the softmax function, while $f_Q$, $f_K$, and $f_V$ represent the query, key, and value mapping functions, respectively. We then apply our Siam-UNet, consisting of two shared-weight UNet branches $\mathbf{U}_1$ and $\mathbf{U}_2$. At each reverse diffusion step, each branch predicts the noise to be removed frame-wisely under the guidance of the task instruction: 
\begin{equation} 
\hat\epsilon^{1:T}_i = \mathbf{U}_i\bigl(\mathrm{MHCA}(\mathbf{z}^{1:T}_i(t), \mathbf{T}(\mathcal{L}))\bigr),
\end{equation}
where $\mathbf{T}$ is the pre-trained text encoder.

During training, we adopt the standard denoising objective~\cite{ddpm} by minimizing the mean squared error (MSE) between the predicted noise and the true noise: \begin{equation} L_{\text{diffusion}}= \sum_i\mathbb{E}_{t,{\epsilon_i^{1:T}},\mathbf{z}_i^{1:T}(0)} \bigl|\bigl|{\epsilon}_i^{1:T} - \hat{\epsilon}_i^{1:T}\bigr|\bigr|^{2}.
\label{eq:diff-loss} \end{equation} This objective encourages the network to accurately predict and remove noise at each reverse step, thus learning a distribution of feasible object-centric motion flows. Finally,
by iteratively correcting the noisy latents, the Siam-UNet converges toward the denoised latent code sequence $\hat{\mathbf{z}}^{1:T}_i(0)$.



\paragraph{Siam-VAE Decoder}

After the Siam-Unet reconstructs the latent code sequences, each of them is passed through a Siamese-structured VAE decoder (Siam-VAE Decoder) frame-wisely as: \begin{equation} \widehat{F}_i=\mathcal{D}\bigl(\hat{\mathbf{z}}_{i}^{1:T}(0)\bigr), \label{eq:decoder} \end{equation} 
where $\mathcal{D}$ is the pre-trained VAE decoder that shares parameters between its two branches, enforcing architectural symmetry with the encoder. Each decoder branch generates a refined 2D motion flow $\widehat{F}_i$ for the corresponding object, capturing fine-grained displacements over time. Together, these two streams of object-centric flows constitute the final output of SFDNet.



Overall, by employing a Siamese encoder–decoder architecture, SFDNet enforces consistent feature extraction for both arms within a joint representation space. Meanwhile, the diffusion process leverages noise injection and language conditioning to yield robust, object-centric motion flows. These 2D flows are then mapped to the 3D execution space and allocated to each arm using the proposed VLM-assisted task assignment strategy, as detailed in the following section.

\subsection{Inference with the VLM-Assisted Spatial-Temporal Task Allocation Strategy}

During the testing stage, we deploy the trained SFDNet directly onto a real-world dual-arm platform. Following AnimateDiff~\cite{animatediff} and Im2Flow2Act~\cite{im2flow2act}, we first randomly initialized the latent code sequences $\mathbf{z}^{1:T}_1(0)$ and $\mathbf{z}^{1:T}_2(0)$ from the standard normal distribution, then SFDNet generates the two-stream object-centric motion flows under the condition of task instruction $\mathcal{L}$ denoted as:
\begin{equation}
    \{\hat{F}_1, \hat{F}_2\}=\mathrm{SFDNet}(\mathbf{z}^{1:T}_1(0),\mathbf{z}^{1:T}_2(0),\mathcal{L}),
\end{equation}
where $\hat{F}_1, \hat{F}_2 \in \mathbb{R}^{3\times T\times H\times W}$ are the predicted motion flows. 
\textcolor{black}{We then follow AVDC~\cite{ko2023learning} to reconstruct 3D trajectories. We first transform 2D points from motion flows in the initial frame into 3D points using the camera’s intrinsic parameters and depth information captured by the RGB-D camera. 
For each subsequent frame, we calculate the following trajectory waypoints by minimizing the reprojection error with randomly initialized transformation matrix $\mathbf{T}_t\in \mathbb{R}^{4\times 4}$ that describes the transformation of 3D waypoint from the first frame to $t^{\text{th}}$ frame. The reprojection error is defined as:
\begin{equation}
    E_t = \sum_{i=1}^{N} \left\| \mathbf{p}_{i,t} - \Pi(\mathbf{T}_t \cdot \mathbf{P}_{i,0}) \right\|^2,
\end{equation}
where $N$ is the number of points within each frame of the 2D flow, $\mathbf{p}_{i,t} = (x_{i,t}, y_{i,t})$ is $i$-th point in frame $t$, $\Pi(\cdot)$ is the projection function mapping 3D points back to the 2D image plane, $\mathbf{P}_{i,0}$ represents the 3D coordinates of the $i$-th point in the first frame. The term $E_t$ represents the reprojection error for frame $t$.
The optimized $\{\mathbf{T}_0, ..., \mathbf{T}_T\}$ can further derive 3D trajectories $\mathrm{Traj}_i \in \mathbb{R}^{T \times 6}$ for two objects ($i \in \{1, 2\}$) that describes $T$ 3D waypoints with coordinates as well as rotation angle. 
These trajectories are refined through post-processing operations, such as smoothing and merging, to ensure coherence. }



The predicted 3D trajectories can be transformed into executable control commands using inverse kinematics, however, resolving workspace constraints and execution timing remains crucial for successful dual-arm coordination. For instance, in a task where two objects must be placed into a basket, executing both trajectories simultaneously without coordination may lead to collisions. Therefore, a task allocation strategy that accounts for both spatial and temporal factors is essential to ensure smooth and effective collaboration between the two manipulators.

\begin{figure*}[h]
    \centering
    \vspace{0.1cm}
    \includegraphics[width=\textwidth]{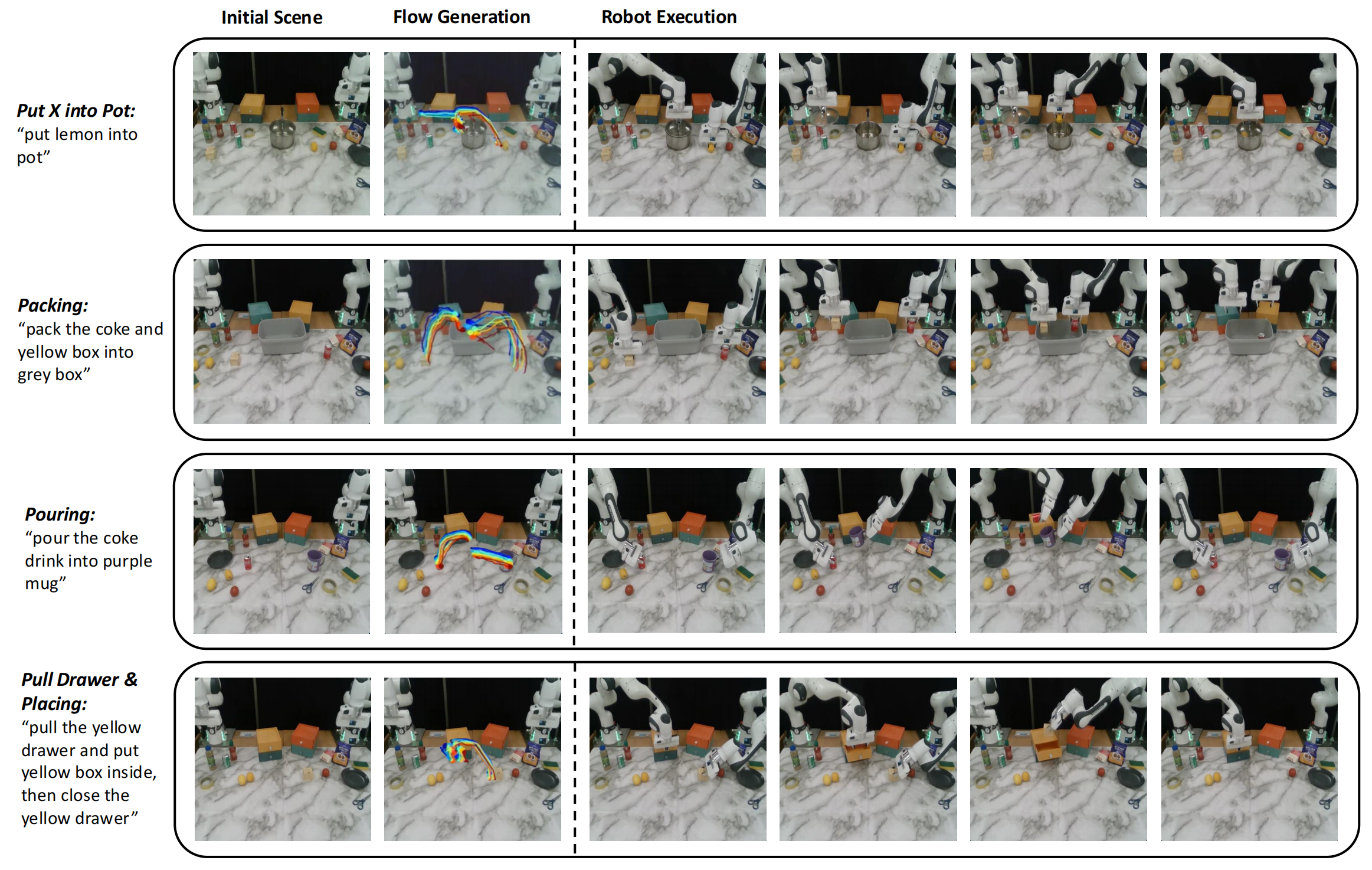}
    \caption{\textbf{Task Examples and Execution Results.} We evaluate our framework on four challenging tasks: (1) Put X into Pot (first row), (2) Packing (second row), (3) Pouring (third row), and (4) Pulling Drawer \& Placing (fourth row). For each task, the figure shows the initial scene, flow generation visualizations, and sequential snapshots of robot execution. 
    }
    \label{fig:exp_traj}
\end{figure*}


To address this, we introduce the VLM-Assisted Spatial-Temporal Task Allocation Strategy, leveraging a pre-trained VLM (GPT-4o~\cite{gpt4o}) for knowledge-driven reasoning. Specifically, instead of executing full-length trajectories simultaneously, \textcolor{black}{
we first overlay the predicted 2D flows on the initial frame $\mathcal{I}_0$. Based on this visualization, the VLM determines the allocation of flows to each arm and decides the segmentation granularity $M_\text{VLM}$, and then calls a pre-designed \texttt{trajectory segmentation} tool $F_\text{seg}$ to partition each predicted flow into smaller 2D segments:
\begin{equation}
\{\mathrm{Seg}_i^1, \mathrm{Seg}_i^2, ..., \mathrm{Seg}_i^{M_\text{VLM}}\} = F_\text{seg}(\hat{F}_i, M_\text{VLM}),
\end{equation}
where each $\mathrm{Seg}_i^m$ represents a fraction of the original flow. The segments are visualized on $\mathcal{I}_0$ with distinct colors and numeric labels, allowing the VLM to jointly analyze all segments and rank them for an optimized execution order, then assigning the open/close control signal for the grippers at the start or end of specific segments. Since each waypoint in the 2D flow corresponds one-to-one with a 3D trajectory waypoint, the selected sub-trajectories can be directly mapped back into 3D space for execution.
}



Since collision and cooperation constraints are inherent to the ranking process, the proposed strategy ensures that (1) no two conflicting sub-trajectories are scheduled simultaneously; (2) dual-arm coordination is naturally optimized, preventing unnecessary delays; (3) execution is both efficient and collision-free without explicit constraint handling.





\section{Experiments}
\label{sec:exp}
To validate the effectiveness of the proposed VLM-SFD framework, we conducted real-world experiments on four challenging dual-arm cooperative tasks, which require accurate spatial-temporal coordination, precise motion control, and effective collision avoidance.
Section~\ref{sec:exp_setup} provides the experimental details, including the real-world setup, task definition, dataset collection, hyper-parameters, baseline methods, and evaluation metrics. 
Section~\ref{sec:exp_analysis} presents the experimental results with corresponding analysis. The comparison results demonstrate that VLM-SFD outperforms all baseline methods and is capable of adapting to tasks in real-world environments with only a few human demonstrations.

\subsection{Experimental Details}
\label{sec:exp_setup}
\begin{figure}[t]
    \centering

    \includegraphics[width=0.8\linewidth]{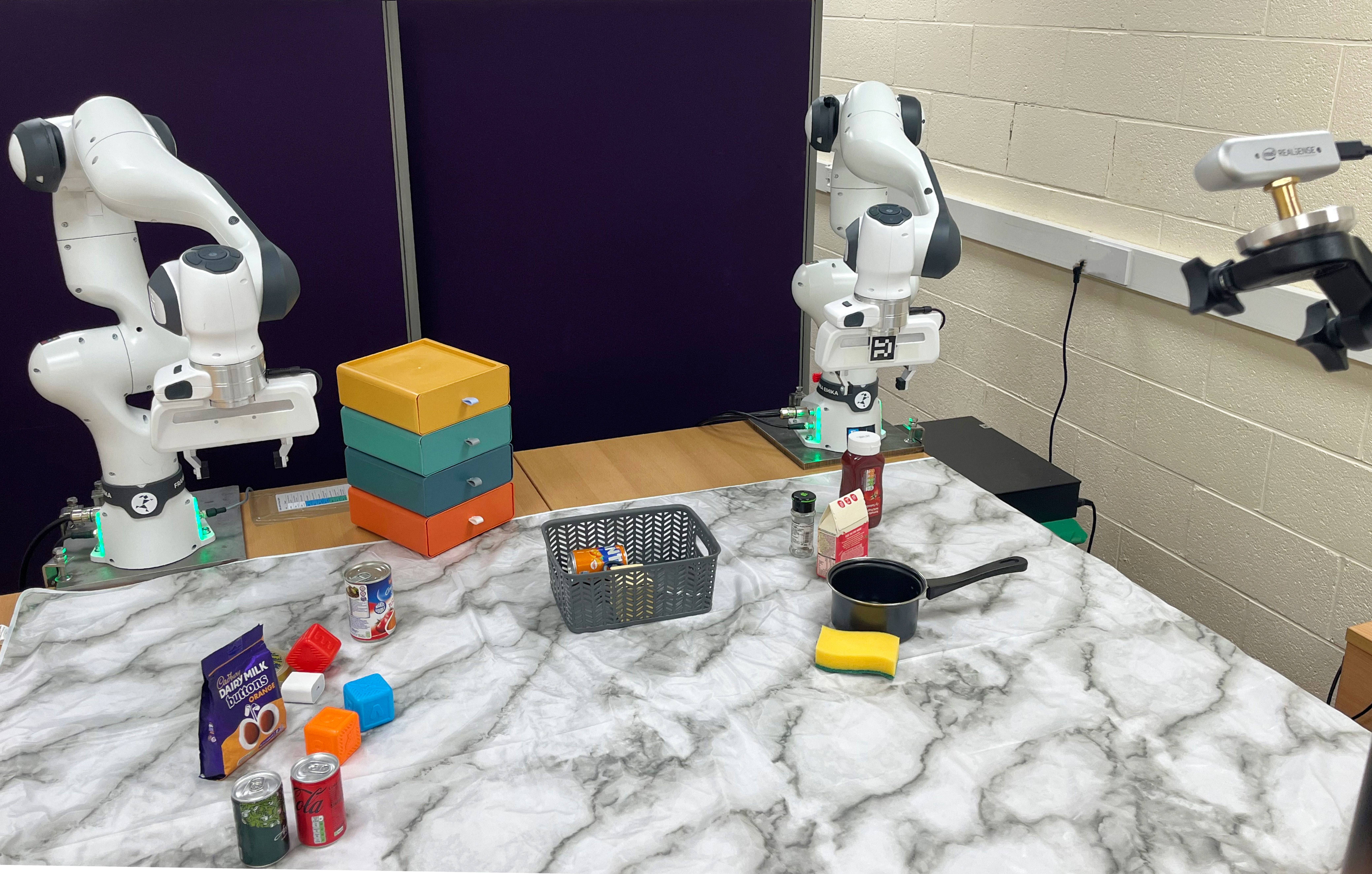}
    \caption{
    Snapshot of the real-world experimental setup: two Franka Research 3 robotic manipulators positioned along with an external Intel RealSense D435i RGB-D camera.
    }
    \label{fig_setup}
\end{figure}

\paragraph{Real-world Setup} As shown in Fig.~\ref{fig_setup}, we employ two \emph{Franka Research 3} robotic manipulators along with an external \emph{Intel RealSense D435i} RGB-D camera mounted in front of the manipulation platform, providing RGB-D images as observations. The positions of all objects are randomly initialized within the robot workspace to ensure task diversity.

\paragraph{Task Definition and Dataset Collection}
We collect training data by human demonstrations. Specifically, we adopt four challenging tasks following the baselines~\cite{zhao2023learning,chi2023diffusion}: 
\begin{itemize}
     \item \textit{Put X into Pot:} The object specified by the instruction should be placed into the pot after removing the lid, and then the lid should be replaced on the pot.
    \item \textit{Packing:} Multiple objects specified by the instruction should be picked up and placed into the box for packing. 
    \item \textit{Pouring:} The can and container (cup or bowl) specified by the instruction should be picked up and moved to the proper work position, then pour the liquid from the can into the container, put down the can and container to their original position. 
    \item \textit{Pulling Drawer \& Placing:} Pull the specified drawer open, pick the specified object and place it into the drawer, then close the drawer.
\end{itemize}
Fig.~\ref{fig:exp_traj} provides examples of each task. These tasks can be categorized into two types of dual-arm collaboration: synchronous (e.g., \textit{Packing}) and asynchronous (e.g., \textit{Put X into Pot}, \textit{Pouring}, and \textit{Pulling Drawer \& Placing}). All manipulated objects are rigid or minimally deformable. For each task, we collect only 10 episodes of human demonstrations involving 3 distinct objects to be manipulated within a cluttered environment.
\textcolor{black}{On average, the demonstration lengths are about 120 frames ($\approx$ 4.8 seconds at 25 FPS).}
The workspace contains numerous distractor items, requiring the robot to accurately identify the object specified by the language instruction while avoiding collisions during manipulation.

\paragraph{Training Details}
Following Im2Flow2Act~\cite{im2flow2act}, we modify and extend AnimateDiff~\cite{animatediff}, an advanced text-to-image diffusion model, into the proposed siamese architecture and fine-tune it on our dataset. 

\textcolor{black}{We adopt the AdamW~\cite{adamw} optimizer with a learning rate of 0.0001 and a weight decay of 0.01. To enable efficient fine-tuning, we employ LoRA~\cite{lora} with a rank of 128. Input images are resized to $224\times224$ before being fed into the model. The training was conducted on a server node equipped with two Intel Xeon Gold 6326 processors and two NVIDIA A100 40GB GPUs.  The total training time was approximately 60 GPU hours on this setup.}

\paragraph{Baseline Methods}
To present the performance of the proposed framework, we conduct the comparison experiments with two advanced dual-arm cooperation baselines:
\begin{enumerate}
    \item \textit{Action Chunking Transformers (ACT)}~\cite{zhao2023learning}. ACT proposes a low-cost imitation learning method, which designs a transformer architecture that combines a Conditional Variational Auto Encoder to predict robot actions from visual inputs.
    \item \textit{Diffusion Policy (DP)}~\cite{chi2023diffusion}. DP formulates robot behavior as a conditional denoising diffusion process within a transformer-based diffusion network, enabling the generation of high-dimensional robot action sequences from images.
\end{enumerate}


\subsection{Experiment Results}
\label{sec:exp_analysis}

\begin{figure}[t]
    \centering   \includegraphics[width=0.9\linewidth]{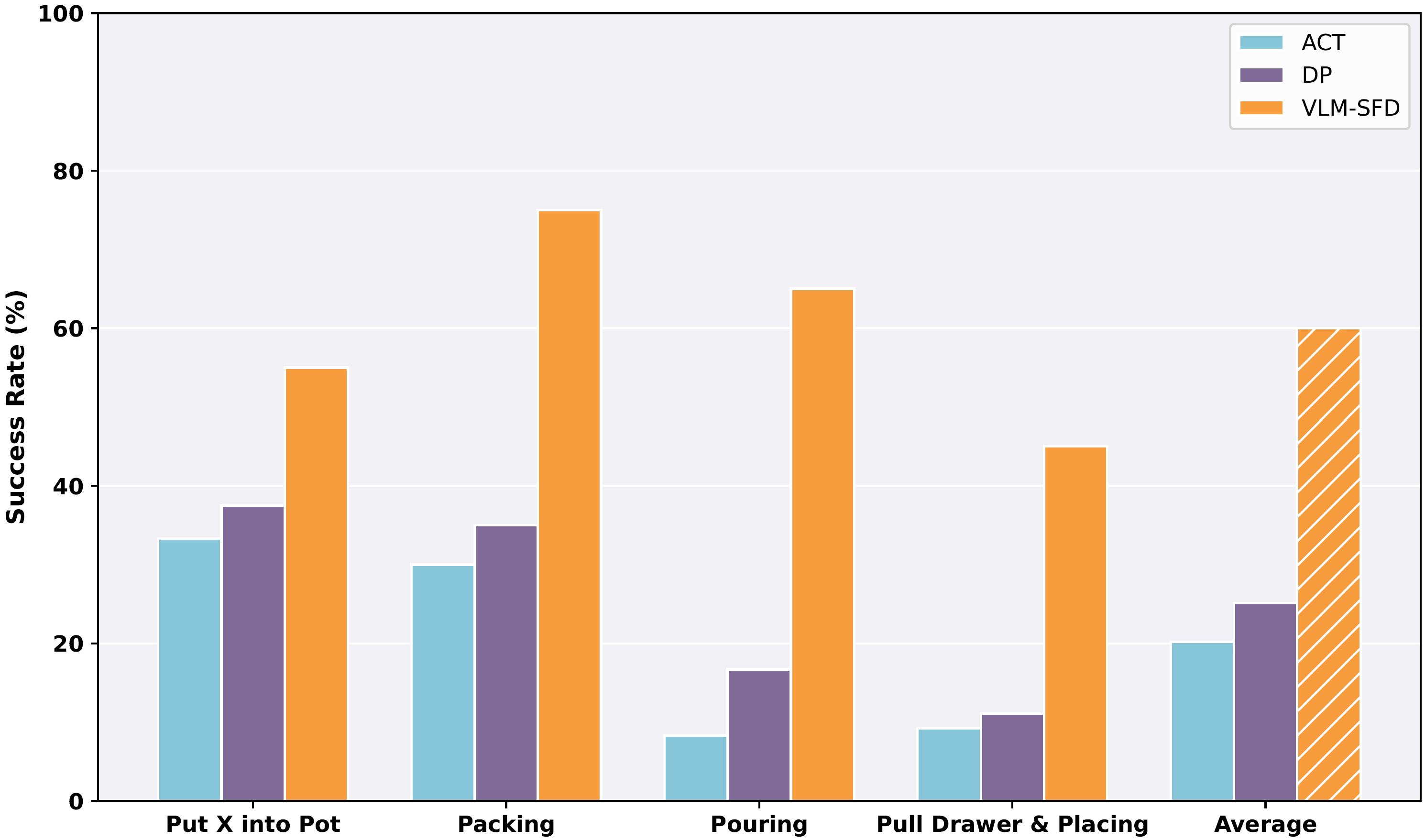}
    \caption{
    Comparison of task success rates (\%) for the proposed VLM-SFD framework vs. baseline methods ACT (blue) and DP (purple) across four dual-arm manipulation tasks. VLM-SFD (orange bars) achieves the highest success rate in every task by a wide margin. All success rates are averaged over 20 repeated trials per task.
    }
    \label{fig_SR}
\end{figure}
The experimental results for real-world tasks are summarized in Fig.~\ref{fig_SR}, with success rates computed from 20 repeated trials per task. The proposed VLM-SFD framework significantly outperforms the ACT and DP baselines across all tasks. On average, VLM-SFD achieves a 60\% success rate, compared to 25\% for DP and 17\% for ACT. This substantial performance margin highlights VLM-SFD’s superior reliability, likely due to its ability to leverage semantic understanding and coordinated planning in cluttered environments.

For the synchronous \emph{Packing} task, which requires precise, simultaneous dual-arm coordination, VLM-SFD reaches 75\% success, far more than DP’s 35\%. This underscores the challenge baseline methods face in generating tightly coupled bimanual actions, whereas VLM-SFD handles them effectively through structured siamese diffusion planning.

In asynchronous tasks involving sequential coordination and task allocation, VLM-SFD consistently achieves the highest success rates. For instance, in the \emph{Pouring} task, it reaches 65\% success, compared to 17\% for DP and only 8\% for ACT. In \emph{Pull Drawer \& Placing}, VLM-SFD achieves 45\%, significantly outperforming DP (11\%) and ACT (9\%). Even in the relatively simpler \emph{Put X into Pot} task, VLM-SFD achieves a 55\% success rate compared to 38\% for DP and 33\% for ACT. These results underscore VLM-SFD’s strength not only in single-step actions but also in complex, multi-stage manipulations. It is largely attributed to the proposed VLM-Assisted Spatial-Temporal Task Allocation Strategy, which effectively reasons over temporal dependencies, spatial constraints, and object semantics to ensure coordinated, efficient, and collision-free execution. 

Furthermore, it is worth noting the advantage in training efficiency of the proposed VLM-SFD framework. Unlike ACT and DP, which require hundreds of binocular video recordings and robot trajectories per task, along with complex post-training like action-level decoding, our framework is trained on a small set of easy-to-collect human demonstrations and can be directly deployed on the real-world robot. 
This comparison demonstrates the efficiency of our method in learning from limited data with low training cost.

\textcolor{black}{To investigate the time cost of the proposed method, we measured the average planning time for flow generation, VLM inference, and the execution time, as shown in Table~\ref{tab:timecost}. These results demonstrate that despite the latency introduced during the planning phase due to the use of the pretrained VLM, the system can still support real-time task execution.}

\begin{table}[t]
\centering
\caption{\textcolor{black}{Planning and Execution Times for Four Tasks}}\label{tab:timecost}
\renewcommand\arraystretch{0.8}
\renewcommand{\arraystretch}{1.2}
\setlength{\tabcolsep}{5pt}
\begin{tabular}{c c c}
\hline
\textbf{Task} & \textbf{Planning Time (s)} & \textbf{Execution Time (s)} \\
\hline
Put X into Pot & 128.4 & 54.3 \\
Packing & 141.9 & 24.8 \\
Pouring & 176.3 & 96.4 \\
Pulling Drawer \& Placing & 146.6 & 64.3 \\
\hline
\end{tabular}
\end{table}


\begin{table}[t]
    \begin{center}    
    \caption{Ablation Studies on Siamese Structure and Task Allocation}
    \label{tab:ablation_putx}
    \begin{tabular}{l c }
        \midrule
        \textbf{Method}                & \textbf{Success Rate (\%)}   \\
        \midrule
        {No Spatial-Temporal Task Allocation}  
                    & 20\%     \\
        {No Siamese structure}  
                    & 40\%     \\
        {Ours(VLM-SFD)}  
                     & 65\%    \\
        \hline
    \end{tabular}
    \end{center}
\end{table}


\subsection{Ablation Study}
We conduct ablation studies to access the contribution of key components in our VLM-SFD framework. The detailed results are summarized in Table~\ref{tab:ablation_putx}. Considering the need for fine-grained control and knowledge-driven spatial reasoning, we select the challenging \textit{pouring} task for 20 repeat tests.

To validate the effectiveness of the proposed VLM-assisted spatial-temporal task allocation strategy, we conduct experiments where both manipulators execute actions simultaneously without it. Under this setting, the model suffers a significant performance degradation, with success rate dropping to 20\%, compared to 65\% achieved by VLM-SFD. Failure cases predominantly arise from temporal misalignment, for example, the can often begins pouring before the cup reaches its target position, causing spillage. These outcomes underscore the importance of explicitly handling temporal coordination within the proposed allocation strategy.

To validate the impact of our Siamese-architecture SFDNet (Siam-VAE encoder, Siam-Unet, and Siam-VAE decoder), we modify the model by using two independent, parallel VAE encoders and U-Nets combined with a shared VAE decoder to generate two separate streams of motion flows. This variant achieves only 40\% success rate, representing a 25\% decrease compared to our Siamese-based design. Analysis of failure cases indicates frequent misalignment between the two manipulators’ targeted positions, primarily due to a lack of shared representation and insufficient coordination in predicting consistent interaction points.


\hltext{
To further examine the influence of the number of demonstrations on task execution, we conducted ablation experiments on the \emph{Pouring} and \emph{Put X into Pot} tasks. As shown in Table~\ref{tab:ablate_num_demos}, for \emph{Pouring}, the success rate increases from 40\% with 5 demos to 65\% with 10 demos, remaining stable with 20 demos. For the \emph{Put X into Pot} task, performance similarly improves from 20\% with 5 demos to 55\% with 10 demos, and with only a marginal gain to 60\% (just one additional successful trial) at 20 demos. These results suggest that around 10 demonstrations are sufficient to achieve stable task execution while balancing training cost and performance.
}

\begin{table}[t]
\centering
\caption{Ablation on number of demonstrations}
\renewcommand{\arraystretch}{1.2}
\setlength{\tabcolsep}{5pt}
\label{tab:ablate_num_demos}
\begin{tabular}{lccc}
\hline
\textbf{Task} & \textbf{5 demos} & \textbf{10 demos} & \textbf{20 demos} \\
\hline
Pouring & 40\% \hltext{(8/20)} & 65\% \hltext{(13/20)} & 65\% \hltext{(13/20)} \\
\hltext{Put X into Pot} & \hltext{20\%}\hltext{(4/20)} & \hltext{55\%}\hltext{(11/20)} & \hltext{60\%}\hltext{(12/20)} \\
\hline
\end{tabular}
\end{table}


\section{Conclusions}
\label{sec:conclusion}
This paper introduces the VLM-Assisted Siamese Flow Diffusion (VLM-SFD) framework for dual-arm cooperative manipulation via imitation learning from only a few human demonstrations. The framework features a Siamese Flow Diffusion Network (SFDNet) that generates two parallel streams of motion flows conditioned on language instructions and visual context, along with a VLM-assisted spatial-temporal task allocation strategy that segments and assigns these motion flows to each manipulator based on temporal dependencies and spatial constraints. VLM-SFD framework empowers to learn complex bimanual tasks efficiently and allows deployment on real-world robotic hardware without fine-tuning. Experimental results demonstrate that VLM-SFD outperforms baselines in four challenging real-world tasks. 

\textbf{Limitations.} 
VLM-SFD relies on a pre-trained vision–language model for task allocation, which introduces non-trivial computational overhead and may affect responsiveness in complex or rapidly changing scenes. \hltext{The current evaluation is limited to controlled settings and has not fully examined robustness under changes in background, lighting, object types, or occlusion. Real-world execution is also influenced by grasping precision, contact stability, and calibration errors, which can affect performance even with accurate flow predictions.} \hltext{As a learning-based imitation approach, the framework lacks the formal safety guarantees of classical motion planning and prioritizes collision avoidance over tight synchronization in some tasks. Future work will enhance environmental and physical robustness, integrate force/tactile feedback, and combine learning-based planning with classical control for safer, more general dual-arm cooperation.}



\printbibliography

\end{document}